\documentclass{article}

\usepackage{arxiv}

\usepackage[utf8]{inputenc} % allow utf-8 input
\usepackage[T1]{fontenc}    % use 8-bit T1 fonts
\usepackage{hyperref}       % hyperlinks
\usepackage{url}            % simple URL typesetting
\usepackage{booktabs}       % professional-quality tables
\usepackage{amsfonts}       % blackboard math symbols
\usepackage{nicefrac}       % compact symbols for 1/2, etc.
\usepackage{microtype}      % microtypography
\usepackage{lipsum}		% Can be removed after putting your text content
\usepackage{graphicx}
\usepackage{amsmath}
\usepackage{float}

\title{Improving Language Generation with Sentence Coherence Objective}

\author{ \hspace{1mm}Ruixiao Sun \\
	Department of Energy Resources Engineering\\
	Stanford University\\
	\texttt{ruixiaos@stanford.edu} \\
	\And
	\hspace{1mm}Jie Yang \\
	Department of Energy Resources Engineering\\
	Stanford University\\
	\texttt{jy0829@stanford.edu} \\
	\And
	\hspace{1mm}Mehrdad Yousefzadeh \\
	Department of Energy Resources Engineering\\
	Stanford University\\
	\texttt{mehrdady@stanford.edu} \\
}

\date{}

\renewcommand{\shorttitle}{Improving Language Generation with Sentence Coherence Objective}

\hypersetup{
pdftitle={Improving Language Generation with Sentence Coherence Objective},
pdfauthor={Ruixiao Sun, Jie Yang, Mehrdad Yousefzadeh},
}

\begin{document}
\maketitle

\begin{abstract}
	Conditional story generation and contextual text continuation have become increasingly popular topics in NLP community. Existing models are often prone to output paragraphs of texts that gradually diverge from the given prompt. Although the generated text may have a reasonable perplexity and diversity, it could easily be identified by human as gibberish.\\
    The goal of our project is to improve the coherence and consistency across sentences in a language-generation model. We aim to solve this issue by first training a sentence pair coherence classifier with GPT-2 pretrained model, and then co-train the GPT-2 language model with this new coherence objective using a method analogous to the REINFORCE algorithm. This fine-tuned language model is able to generate lengthy paragraph conditioned on a given topic without diverging too much. The simplicity of this model allows it to be applicable to a variety of underlying language model architecture since it only modifies the final layer of the pre-trained model.
\end{abstract}

% keywords can be removed
% \keywords{First keyword \and Second keyword \and More}

\section{Introduction}
Conditional story generation and contextual text continuation have become increasingly popular topics in NLP community. The problem setup usually consists of a user-provided prompt sentence, and the story generation model is required to generate text that revolve around the prompt's topic. This problem is different from language modeling in the sense that we are not only interested in predicting the next word given the previous words, but also the coherence between the generated text and the given prompt. 

\begin{figure}[H]
	\centering
	\includegraphics[width = 0.9 \linewidth]{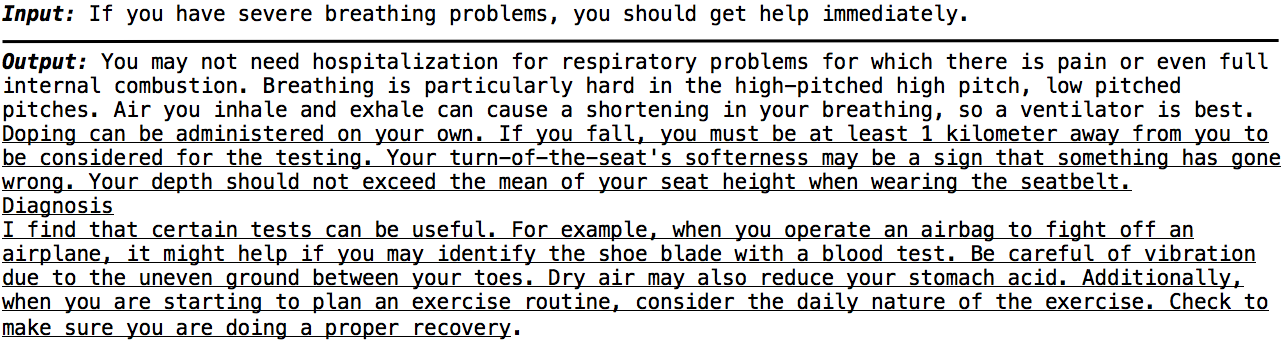} 
	\caption{Example of a GPT-2 text generation output, underlined text shows sentences drifting off topic}
	\label{gpt2_generate}
\end{figure}

Large pretrained transformer-based models such as BERT \cite{devlin2018bert} and GPT-2 \cite{radford2019language} have reached state-of-the-art result on many common NLP tasks. GPT-2 model is a casual language model that predicts the next word conditioned on all the previous words. Naturally, one would be tempted to use this model to make conditional story generation. Indeed, the authors showed some good examples in the original paper. However, from our experience, the generated text still exhibit some degree of incoherence, that is, drifting off the topic for long paragraph of text. See Fig.\ref{gpt2_generate} for an example.

Our idea is to leverage the existing architecture of the pretrained transformer-based language model such as GPT-2, and turn it into a conditional story generation model without having too much modification. Compared to other models specifically targeting at story generation \cite{fan2018hierarchical}, our model is faster to training because of the pretraining, and the contextual representation allows it to model longer term dependency. 

%%%%%%%%%%%%%%%%%%%%%%%%%%%%%%%%
\section{Related Work}
Seq2seq RNN models have been traditionally used to generate conditional stories \cite{roemmele2016writing}. More recently, a hierarchical architecture of RNN models proposed by Fan, et al \cite{fan2018hierarchical} has achieved great performance on conditional story telling by first generate a premise and then the story. The advance of large-scale pretrained transformer models such as GPT-2 \cite{radford2019language} has revolutionized lots of NLP tasks including language modeling. Its ability to extract long-distance contextual information leads to its great success in generating text with great semantic and syntactic coherence. GPT-2 has been applied to conditional language generation and image captioned in an encoder-agnostic setting by Ziegler, et al\cite{ziegler2019encoder}. In this paper, we would like our language model to conform to a new objective that takes into account the correlation between generated text and the prompt text. But this auxiliary objective is non-deterministic with respect to the model parameters since the story is randomly generated from a distribution that depends on the model parameters. 

For models where the dependency between the objective function and model parameters are non-deterministic, or models where the objective is of discrete value thus non-differentiable, researchers often resorts to techniques used in Reinforcement Learning community. A specific set of algorithms are categorized as REINFORCE algorithm \cite{williams1992simple}.
Co-training a generation model with auxiliary discrete objective functions has been applied to the field of music generation using Reinforcement Learning techniques, in which the objective consists of a music generation objective and a reward based on music theory \cite{jaques2016generating}.
REINFORCE algorithm has been applied in a similar context to text auto-completion system by Lee, et al\cite{lee2019learning}.
%%%%%%%%%%%%%%%%%%%%%%%%%%%%%%%%%

\section{Approach}
To train a language model with an additional coherence objective, we have to first make predictions or evaluate whether two sentences followed a coherent topic. In the first step, we are training a coherence classifier on top of GPT-2 model. This allows us to (1) generate a sentence embedding for the prompt sentence (2) rank the coherence between two sentences. And in the second step, we co-train a modified language model that takes in the prompt sentence embedding as a "shortcut" or "guidance", with an additional coherence objective forcing the generated text to be aligned with the prompt's topic.

\textbf{Sentence Coherence Prediction:}
 We implemented our own coherence prediction model using GPT-2 as our underlying pretrained transformer and fine-tune it to predict the sentence coherence.

\begin{figure}[H]
	\centering
	\includegraphics[width = 0.7 \linewidth]{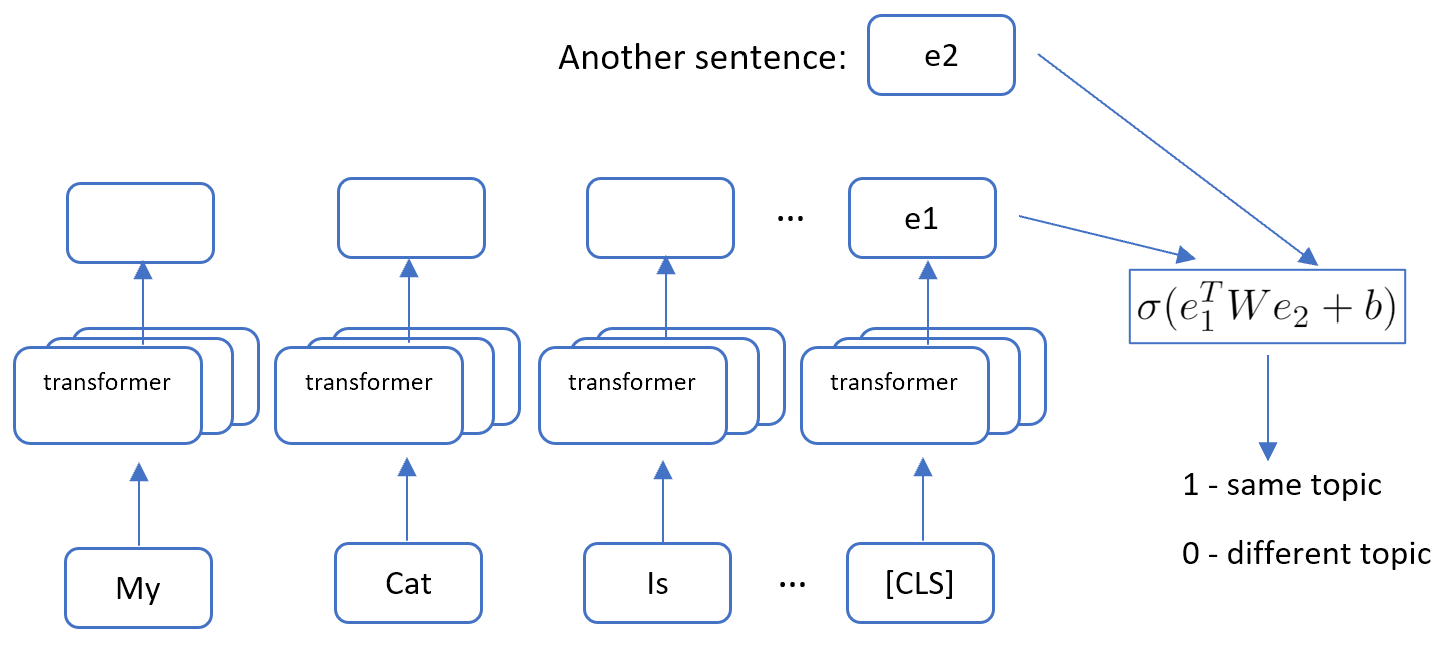} 
	\caption{Architecture of the coherence-prediction model}
	\label{illustration}
\end{figure}

The basic idea is that, if two sentences appear in the same paragraph in our training set, they must have a common underlying topic. The model proposed here will be analogous to the sentence embedding approach first proposed by Kiros, et al \cite{kiros2015skip} and extended by Oord, et al \cite{oord2018representation}. We encode each sentence by adding [CLS] token to the last position, and feed the hidden state of this token to a double dot-product regression model. The final output is from a logistic regression predicting if the two sentences come from the same paragraph or not. 
The binary classification problem is formulated as the following:

\begin{equation}
    p_{coherence}(e_1, e_2)=\sigma(e_1^TW e_2 + b) 
\end{equation}

where $e_1$ and $e_2$ are embeddings for sentence 1 and setence 2 respectively and $\sigma$ is the sigmoid function. Aside from fine-tuning the parameters of the transformer model, we have new trainable parameters, including: weight matrix $W$, bias $b$, weights for the [CLS] token. The architecture is shown in Fig.\ref{illustration}

The training step involves:
\begin{itemize}
    \item Preprocess the dataset. One dataset contains adjacent sentences from the same paragraph, and one dataset contains random sentences from the entire training set.
    \item Load GPT-2 model with pretrained weights for known word pieces. Add an additional token [CLS] with randomly initialized weights. The hidden state of this [CLS] token will be considered as the sentence embedding. Then we create another double dot-product classifier with random initial weights, which takes as input the two sentence embeddings and predict if they should be adjacent.
    \item Train the model with a mixed objective: $L = L_{lm} + \lambda L_{coherence}$. Where $L_{lm}$ is the casual language model objective with cross-entropy loss, i.e. $L_{lm} = -log\frac{exp(x[label])}{\sum_i exp(x[i])}$. And $L_{coherence}$ is the binary cross-entropy loss for coherence prediction, which is $L_{coherence} = y log(\sigma(x)) + (1-y) log(1-\sigma(x))$. Label 1 indicates that two sentences come from the same paragraph. $\lambda$ is a weighting factor for the coherence objective. For each iteration we have one positive sample and one negative sample picked randomly from the training set.
    \item Evaluate the model on validation set. We have 50\% positive samples and 50\% negative samples randomly chosen from validation set. The ultimate criterion is the percentage of accuracy on the test set.
\end{itemize}

Note that if we concatenate two sentences together, separated by [SEP] token, we could take the last hidden state of [CLS] token and feed it into logistic regression. This proves to be highly efficient and accurate because the two sentences could attend to each other, which is recommended in the original OpenAI GPT paper \cite{radford2018improving}. However, this approach is infeasible for our problem, because in language generation model, each word is generated on the fly as opposed to given a-priori. We will compare the prediction accuracy of the sentence embedding approach with this cross-attention approach in the following section.

\textbf{Text Generation Model: } In order to generate the text we used Conditional text generation using the auto-regressive models of the GPT-2\cite{radford2019language}. GPT-2 is a large transformer-based language model trained on a dataset of 8 million web pages. GPT-2 is trained with a casual language modeling objective, that is: given all of the previous words within some text, predict the probability of the next word\cite{radford2018improving}. The probability of a sentence in this sense can be written as: $p(x) =  \Pi_{i=1}^n p(s_i|s_1, s_2, ..., s_{i-1})$   

The original implementation of the conditioned text generation model we used comes from pretrained GPT-2 small model from "Huggingface Transformers" (https://huggingface.co/transformers/model\_doc/gpt2.html). It was trained on 12-layer, 768-hidden, 12-heads, 117M parameters. We fine-tuned the model on Wikipedia dataset (https://blog.einstein.ai/the-wikitext-long-term-dependency-language-modeling-dataset/)

Co-training objective: in a casual language model, the next token is conditionally dependent on all previous tokens, which is what the original GPT and GPT-2 model does. In our model, however, we also want our model to be more conforming to the given prompt without diverting to another topic. To achieve this, we added a feed-forward neural network for each hidden state before deciding the output token. The layer takes as input the concatenation of the original hidden state and the sentence embedding, and output a hidden state of the same size as the original hidden state. The scores of each token are then calculated by multiply it to the vocabulary embedding matrix. Written in terms of equations:

\begin{align}
    H_1 = W_1[e_p ; h_i] + b_1 \nonumber\\
    H_2 = W_2 \cdot tanh(H_1) + b_2 \nonumber\\
    Scores = V H_2
\end{align}

\begin{figure}[H]
	\centering
	\includegraphics[width = 0.8 \linewidth]{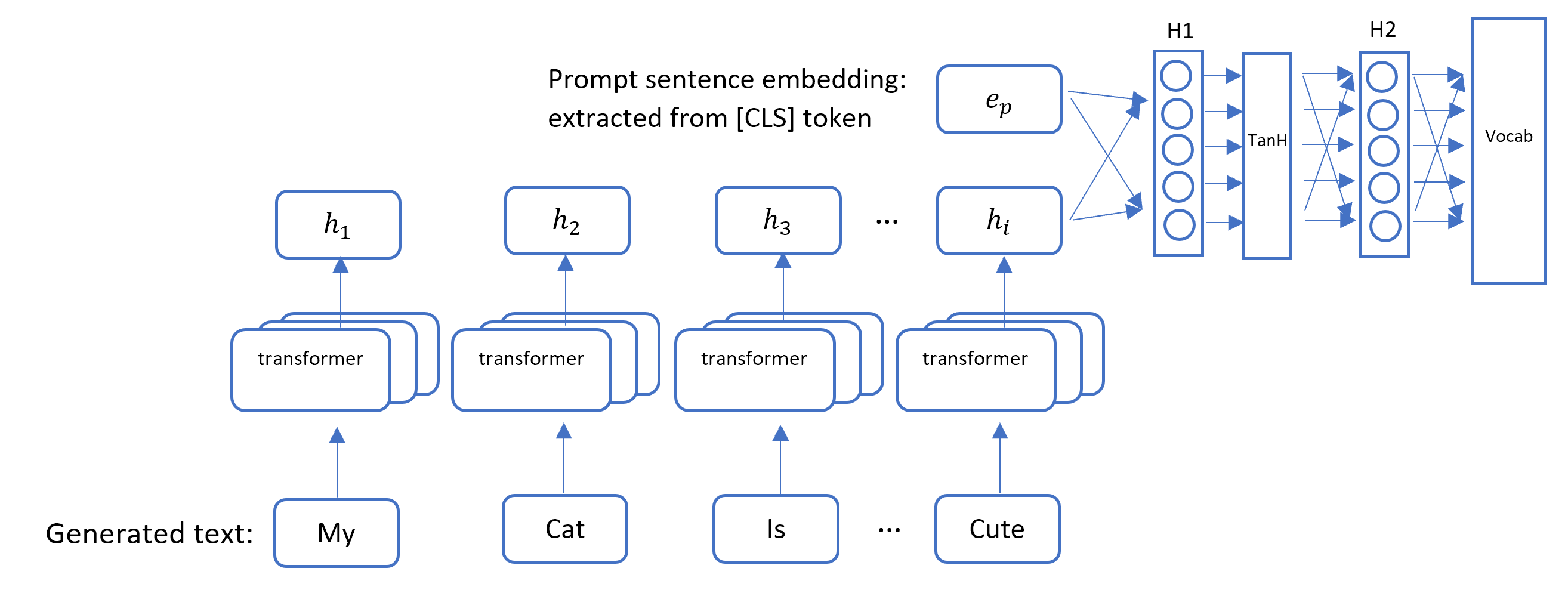} 
	\caption{Architecture of the text-generation model}
	\label{lm}
\end{figure}

See figure.\ref{lm} for an illustration. The size of the layer $H_2$ is the same as the output hidden state $h_i$, because in this way, the vocabulary embedding matrix $V$ can be initialized as the input embedding matrix without any modification. The prompt sentence embedding here is acting as a "shortcut" for the next-token prediction, with the hope that this will force the next token to be more aligned to the given prompt. In the result section, we also compared this architecture with one with linear projection instead of nonlinear layers. And it showed that linear projection could lead to degenerated text.

Correspondingly, we have an enhanced language modeling objective. Aside from the cross-entropy loss for next-token-prediction, there is another loss measuring the coherence and similarity between the generated text and the prompt text. Since we used logistic regression in the previous task to classify if two sentences are from the same topic, a natural choice for this loss is the negative of the logit for that classification model. The total loss can be written as $L = L_{lm} + \lambda_2 L_{similarity}$. The same as before, $L_{lm}$ is the casual language model objective with cross-entropy loss, i.e. $L_{lm} = -log\frac{exp(x[label])}{\sum_i exp(x[i])}$. Note that the language model loss $L_{lm}$ is per-word based and the similarity loss $L_{similarity}$ is per-sentence based. We calculate the similarity loss after the model generated a complete paragraph.

During this step, the only parameters we are training are $W_1, b_1, W_2, b_2$. The functional relationship between $L_{similarity}$ and the parameters are indirect: we have to generate a whole sentence before plugging it into the coherence prediction model. But the text is sampled from a distribution instead of a deterministic relationship with respect to the model parameters. Considering this difficulty, we decided to use a variant of the REINFORCE algorithm \cite{williams1992simple} to approximate the gradient w.r.t. the parameters. Written in terms of equations:

\begin{equation}
\begin{aligned}
    L_{similarity} &= -E[\sum CrossEntropy  \cdot (R-\bar{R})] \\
    \nabla L_{similarity} &= -E[\sum \nabla CrossEntropy  \cdot (R-\bar{R})] \\
    R &= e^TW e_p + b \\
    \bar{R} &= \alpha R_{current} + (1-\alpha) R_{previous}
\end{aligned}
\label{REINFORCE}
\end{equation}

where $R$ is the reward function calculated for each generated text, and $\bar{R}$ is a running exponential average of previous rewards with parameter $\alpha$. The term $\sum \nabla CrossEntropy$ is the sum of the cross-entropy loss for each of the generated token. The true lables for this cross-entropy loss is set to be the generated text itself. Finally, we perform stochastic gradient descent with this approximated gradient, using a weighting factor $\lambda_2$.  The rationale behind this is: if a piece of generated text received high rewards, meaning a coherent sentence w.r.t. the prompt, we will update the weights such that it generate more sentences like this. And vice versa. Note that the gradient calculation using REINFORCE is noisy and does not converge as quickly as normal gradient descent.

The training procedure involves the following steps
\begin{itemize}
    \item Preprocess the dataset. Segment the corpus into pieces of sentences of length about 200 tokens. 
    \item Load GPT-2 model with pretrained weights as well as the trained binary classifier. 
    \item Co-train the language model with a coherence objective: $L = L_{lm} + \lambda_2 L_{similarity}$. 
    For each training step, we first train it using casual language model with cross-entropy loss $L_{lm}$. Secondly, we take the next sentence as the prompt sentence, let the model generate a complete paragraph of text. We then calculate the coherence between the prompt sentence and generated text using our binary classifier. The score is fed into the REINFORCE algorithm \ref{REINFORCE} to calculate the approximated gradient.
\end{itemize}

\textbf{Baselines:}
Since our tasks are divided into two parts: coherence prediction and text generation, we have different baselines for them. For the coherence prediction problem, we compare sentence embedding approach with cross-attention approach and human evaluation. For text generation, the baseline is the vanilla GPT-2 model without any modification. We are evaluating the consistency of generated text manually to see if there is any jump in topic.

\section{Experiments}
The training and evaluation are divided into two steps: sentence coherence model and text generation model.
Both of them shared the same data set: Wikipedia dataset \cite{merity2016wikitext} is applied for training and fine-tuning the model. Currently we trained the model on raw character level dataset WikiText-2. 

\subsection{Sentence Coherence Model}
\textbf{Data}: To train the coherence classification model, we have to randomly choose positive samples and negative samples from the data set. Positive samples come from the same paragraph as the anchor text, negative samples are randomly picked from the entire training set.

\textbf{Evaluation method}: We evaluate the accuracy in percentage on the test set. The accuracy is defined as the number of correctly predicted labels divided by the total number of input pairs (true positive). As describe in the previous section, we will compare the prediction accuracy of the sentence embedding approach with this cross-attention approach. We also compared the results with human performance using 50 sentences, in which 25 of them are positive sample and 25 negative. The human evaluator correctly predicted 46 samples, as well as 3 false positive samples and 1 false negative samples.  

\textbf{Experimental details}: In our experiment, hyperparameters are set as follows: weighting factor of the coherence objective is 0.05, the learning rate of Adam optimizer being 5e-5, the adam epsilon being 1e-8, number of training epoch being 4, batch size per GPU/CPU for training/evaluation being 2.

\begin{figure}[H]
	\centering
	\includegraphics[width = 0.35 \linewidth]{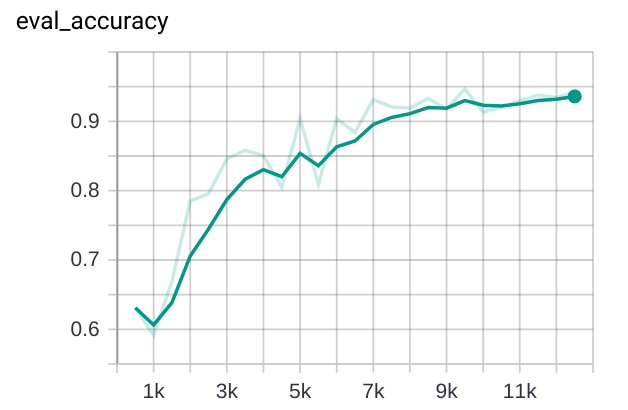} 
	\includegraphics[width = 0.35 \linewidth]{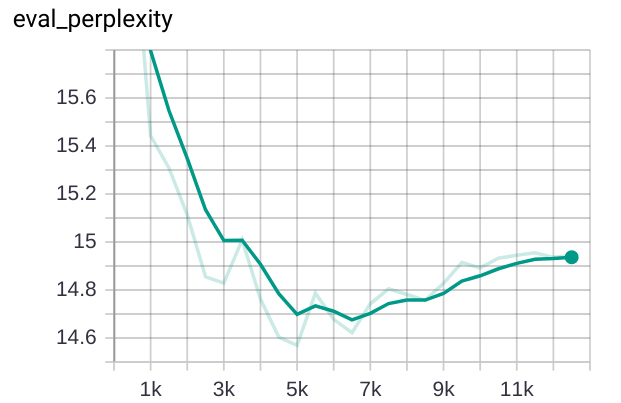}
	\caption{Coherence prediction accuracy and perplexity vs. iteration number on the validation set, visualized using TensorBoard with smoothing=0.6}
	\label{gpt2_example}
\end{figure}

\textbf{Results}: The accuracy results are shown in Table.\ref{tb} and the training process shown in Fig.\ref{gpt2_example}. We can see that the coherence classification model achieves human-level performance, and the performance is close the cross-attention classifier. Since the accuracy is near 95\%, we can assume that the classifier is a good indicator of the coherence between two sentences, and the underlying sentence embedding is a decent summary of its content.  

\begin{table}[H]
\centering
\begin{tabular}{@{}cc@{}}
\toprule
Methods           & Accuracy (\%) \\ \midrule
Sentence embedding & 94.21         \\
Cross-attention   & 94.52         \\
Human performance & 92              \\ \bottomrule
\end{tabular}
\caption{Accuracy of coherence prediction on the test set}
\label{tb}
\end{table}

\subsection{Text generation model}
\textbf{Data:} We take random sentences of length 300-500 from the training set to train the language model. As for the auxiliary coherence objective, we take a random piece of text of length 100 tokens from the training set as the prompt sentence, generate a 500-token long passage of text, and then do parameter update using REINFORCE algorithm.

\textbf{Evaluation methods:} We will manually check the consistency of the output. An example of text generation output with a prompt input is presented in Fig.\ref{gpt2_generate}, from which we notice the generated sentences becomes incoherent (the underline part) to the input sentence.

\textbf{Experimental details:} hyperparameters are set as follows: weighting factor of the auxiliary objective is 1, the learning rate of Adam optimizer being 5e-5, the adam epsilon being 1e-8, number of training epoch being 4, batch size per GPU/CPU for training/evaluation being 4. The auxiliary objective is active after the second epoch, this is because for the first epoch, the language model is not yet tuned, resulting in an unphysical coherence objective.

\begin{figure}[H]
	\centering
	\includegraphics[width = 0.35 \linewidth]{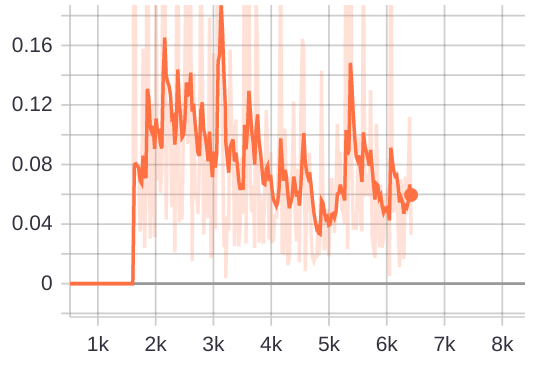} 
	\includegraphics[width = 0.35 \linewidth]{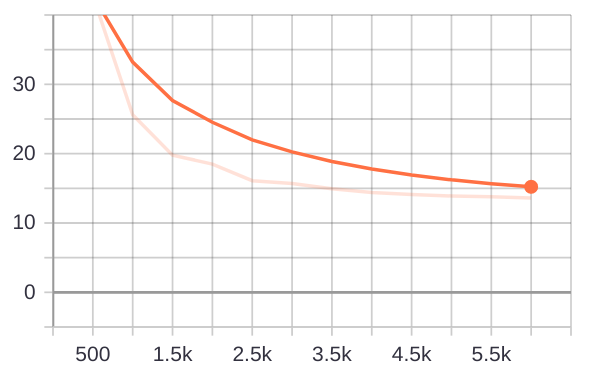}
	\caption{Left: Coherence loss vs. iteration number during training, Right: perplexity value on the validation set during training, visualized using TensorBoard with smoothing=0.8}
	\label{text_gen}
\end{figure}

\textbf{Results:}
The training procedure is illustrated in Fig.\ref{text_gen}, in which the coherence loss is calculated based on Eq.\ref{REINFORCE}. Lower value means better consistency. As can be seen, the coherence loss is very noisy, because: (1) the generated text is randomly sampled (2) the REINFORCE algorithm computed a noisy approximation of the gradient. But the general trend indicates that the coherence loss goes down over time. The perplexity value on the validation set converges to 15. 

To compare with the original GPT-2 model, we can look at an example of a generated piece of text from our model, using the same input prompt as Fig.\ref{gpt2_generate}. We argue that the output from our model is significantly more consistent, and more closely aligned to the prompt sentence's underlying topic. To see more examples and comparison, please refer to the appendix.
\begin{figure}[H]
	\centering
	\includegraphics[width = 0.9 \linewidth]{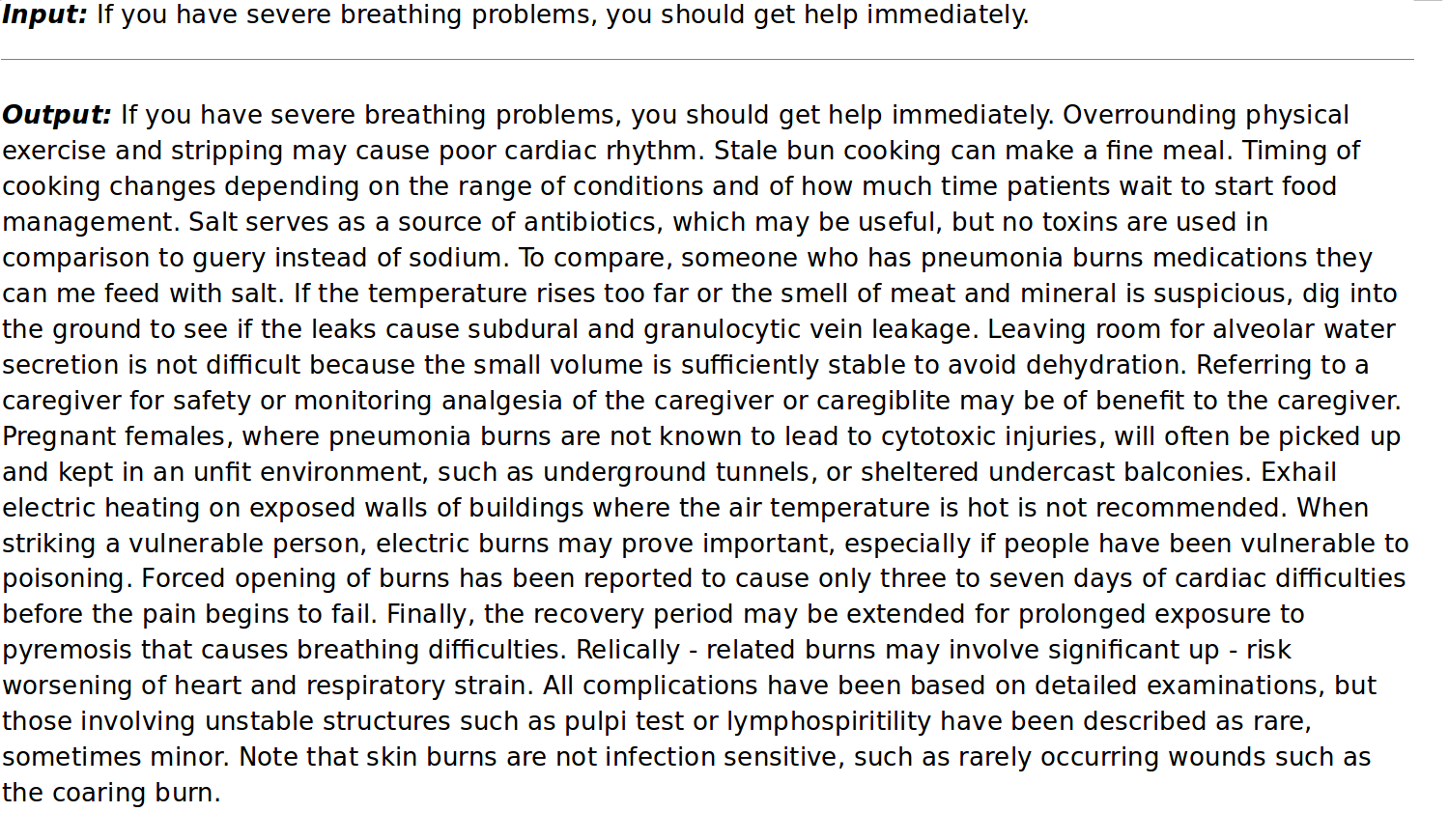} 
	\caption{Example of a text generation output using our model}
	\label{forced}
\end{figure}

\section{Analysis}
Our language generation model co-trained with an auxiliary coherence objective is able to produce long paragraph of text that is aligned with the underlying topic of the prompt sentence. Pure language models such as GPT-2 will prone to drift off the topic over time. We have tried linear projection and nonlinear transformation of the prompt sentence embedding onto the next word prediction, results indicate that linear transformation often leads to degenerated text with repetitions, whereas nonlinear models have great performance (See appendix for an example). The reinforcement learning loss does not monotonically decrease with time. But over a longer period, the average loss does get lower.

Although the generated text from our model is able to stick to the prompt sentence's topic, the wording and grammar are sometimes not as natural as the pure GPT-2 language model. We believe this is a price to pay when we force each generated word to align with a given topic.

\section{Conclusion}
In this work, we have introduced the problem of text incoherence when applying a pure language model to a conditional story-generation problem. We proposed a new conditional language generation model following two steps. The first step involves training a sentence coherence classifier, which also enables us to generate a sentence embedding for each prompt sentence. The second step involves co-training a conditional language model with an auxiliary coherence objective. This new objective is optimized using REINFORCE algorithm. After some experiments, we found out that our model is able to produce long text without diverging from the main topic. However, the wording and grammar are sometimes not as natural as the pure GPT-2 language model. Future work will be focused on how to improve the fluency of the output while still maintaining the coherence between generated text and the prompt.

\bibliographystyle{unsrt}


\appendix

\section{Appendix}
In the following, we are going to show the conditional story generated from 3 models: our proposed model, original GPT-2 model, and a model with linear projection from sentence embedding to the output (see experiment section for more details). All 3 examples followed the same prompt sentence, which described a new gaming console called Playstation 5.

Similar to out previous findings, the proposed model is able to generate long passage of text without drifting off the topic, whereas GPT-2 model gradually diverge from the given prompt topic. Linear model does not have a good performance and is prone to stuck in self-repetition.

\begin{figure}[h]
	\centering
	\includegraphics[width = 0.9 \linewidth]{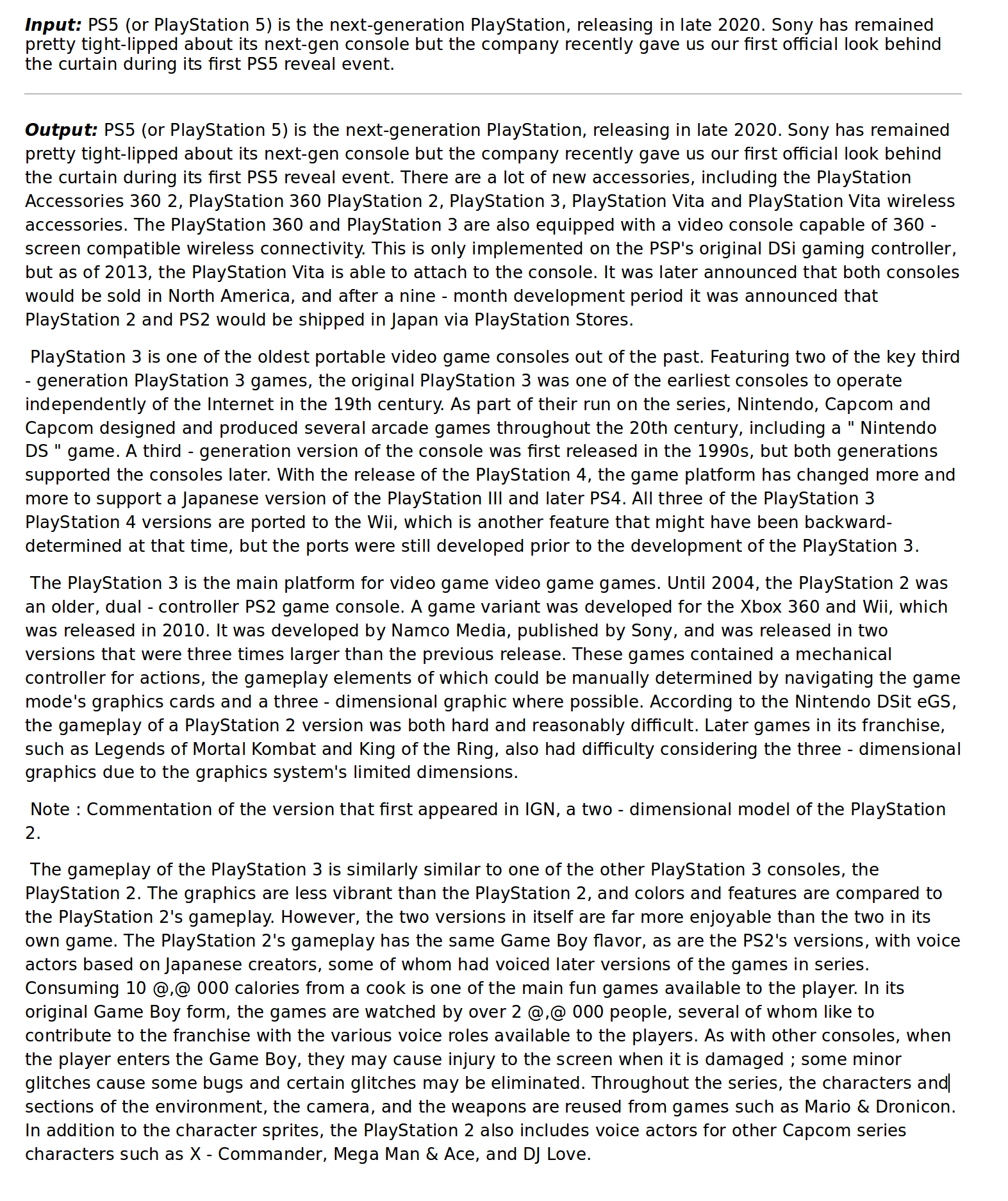} 
	\caption{Example of a text generated using our proposed model}
	\label{forced}
\end{figure}

\begin{figure}[h]
	\centering
	\includegraphics[width = 0.9 \linewidth]{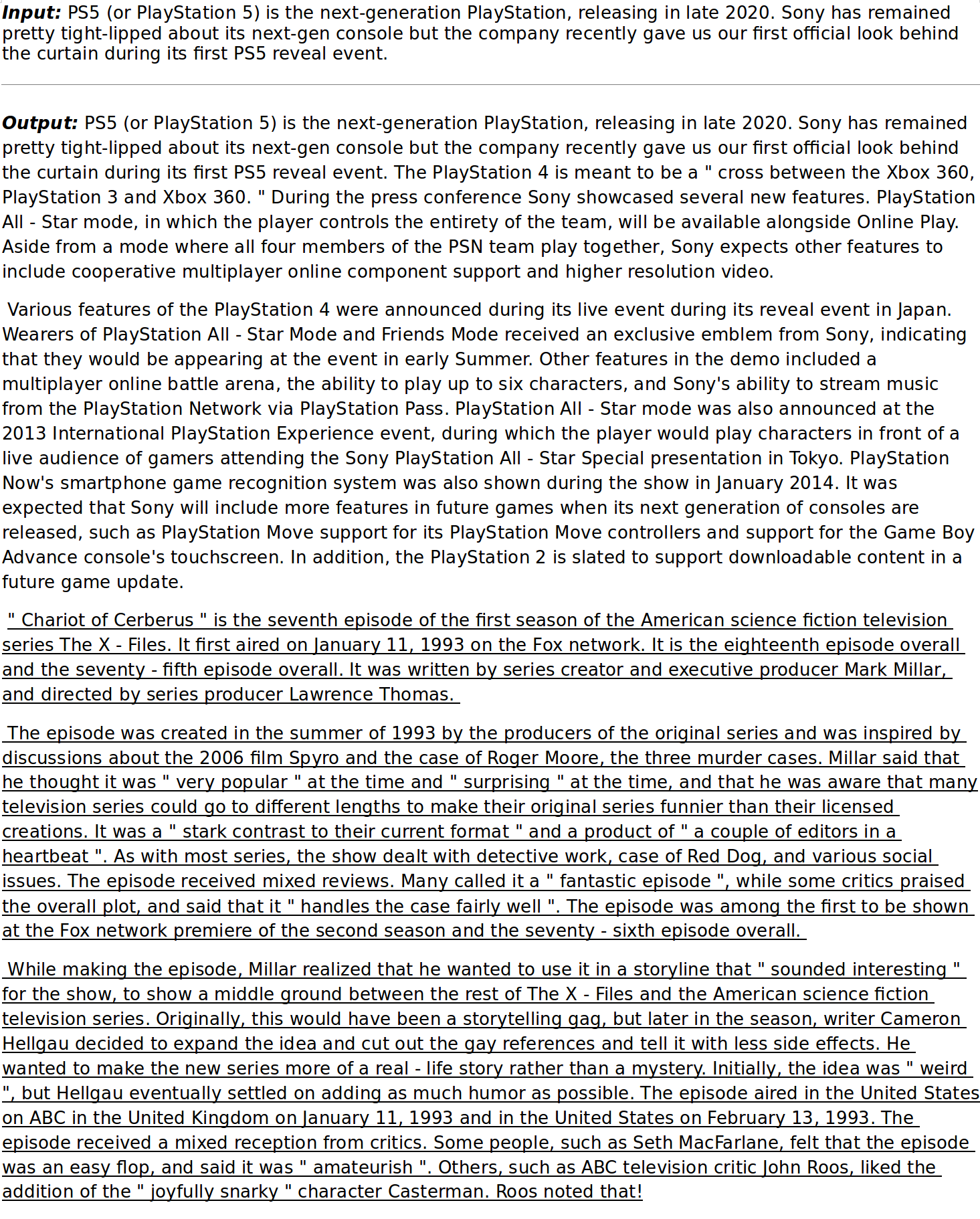} 
	\caption{Example of a text generated from the original GPT-2 model. Underlined text shows the portion of text diverging from the given topic}
	\label{forced}
\end{figure}

\begin{figure}[h]
	\centering
	\includegraphics[width = 0.9 \linewidth]{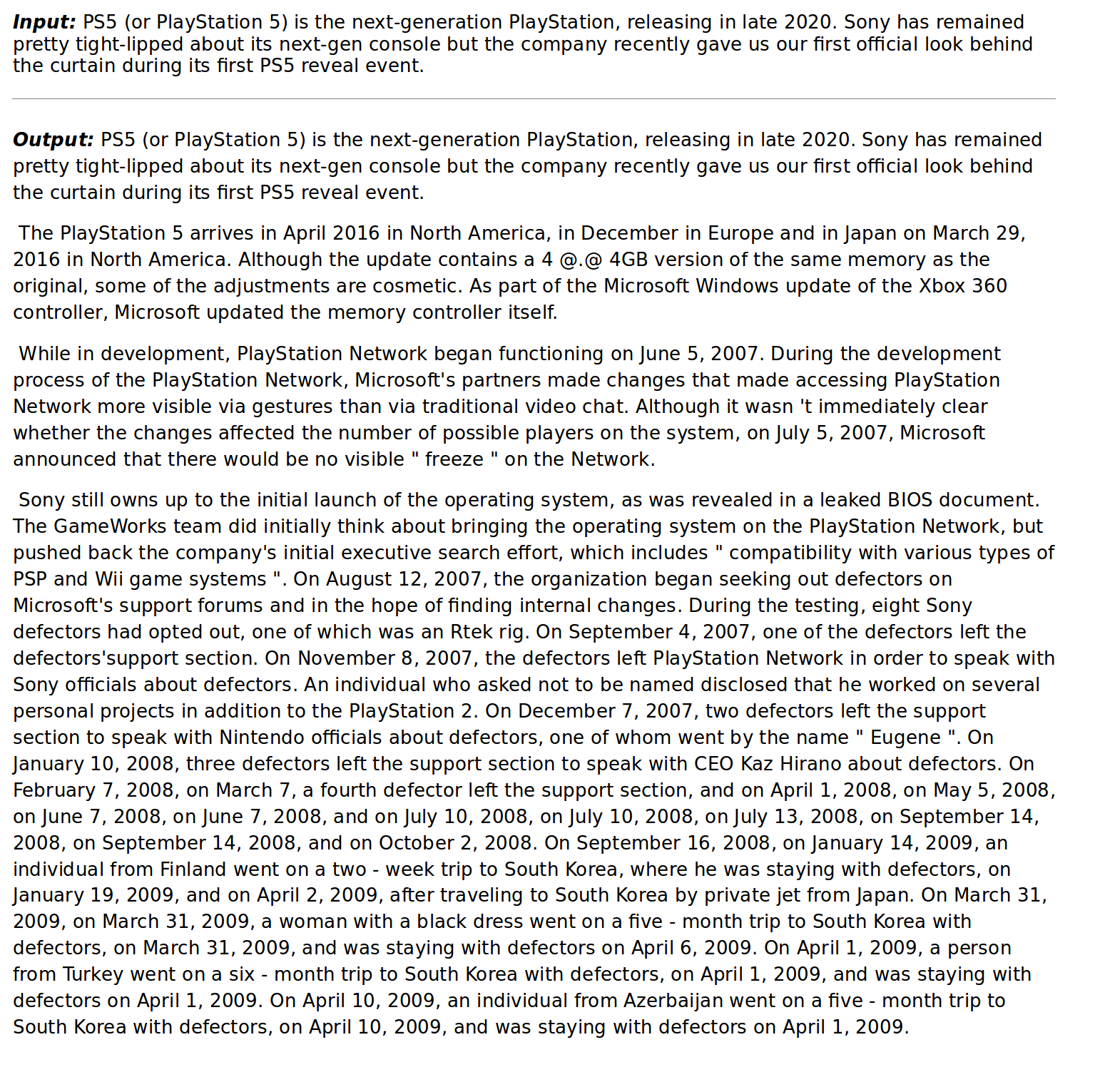} 
	\caption{Example of a text generation output using a linear model. Notice the repetition in the last paragraph}
	\label{forced}
\end{figure}

\end{document}